\documentclass[a4paper]{article}

\usepackage{INTERSPEECH2022}
\usepackage{microtype}
\usepackage{url}

\usepackage{hyperref}

\title{Attention Enhanced Citrinet for Speech Recognition\thanks{Accepted by InterSpeech 2022.}}

\name{Xianchao Wu}
\address{NVIDIA}
\email{xianchaow@nvidia.com}

\begin{document}

\maketitle
\begin{abstract}
  Citrinet is an end-to-end convolutional Connectionist Temporal Classification (CTC) based automatic speech recognition (ASR) model. To capture local and global contextual information, 1D time-channel separable convolutions combined with sub-word encoding and squeeze-and-excitation (SE) are used in Citrinet, making the whole architecture to be as deep as including 23 blocks with 235 convolution layers and 46 linear layers. This pure convolutional and deep architecture makes Critrinet relatively slow at convergence. In this paper, we propose to introduce multi-head attentions together with feed-forward networks in the convolution module in Citrinet blocks while keeping the SE module and residual module unchanged. For speeding up, we remove 8 convolution layers in each attention-enhanced Citrinet block and reduce 23 blocks to 13. Experiments on the Japanese CSJ-500h and Magic-1600h dataset show that the attention-enhanced Citrinet with less layers and blocks and converges faster with lower character error rates than (1) Citrinet with 80\% training time and (2) Conformer with 40\% training time and 29.8\% model size.
  

\end{abstract}

\section{Introduction}





\begin{figure}[t]
  \centering
  \includegraphics[width=6.5cm]{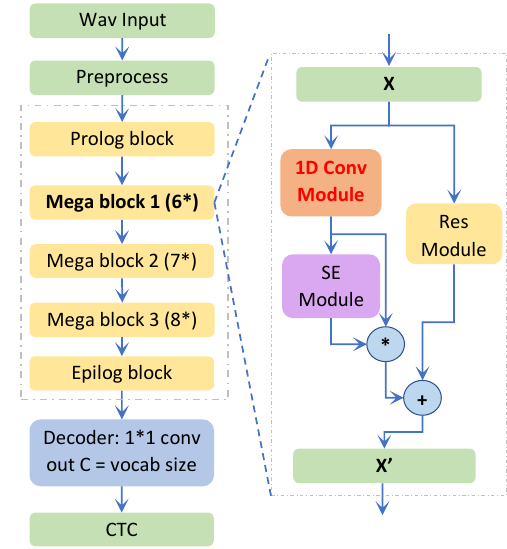}
  \caption{The end-to-end architecture of Citrinet and its major building block - ``1D time-channel separable convolution and SE-attached'' Jasper Block.}
  \label{fig:jasper_block_with_se_in_citrinet}
\end{figure}

Convolutions with predefined kernel sizes capturing local contexts have been successfully utilized in end-to-end ASR systems. JasperNet \cite{Li2019JasperAE}, QuartzNet \cite{Kriman2020QuartznetDA}, ContextNet \cite{Han2020ContextNetIC}, and Citrinet \cite{Majumdar2021CitrinetCT} follow this direction and use pure convolution networks to capture the interactive context information of input wave spectrogram. Among these models, Citrinet (Figure \ref{fig:jasper_block_with_se_in_citrinet}) has achieved top-level accuracy. Citrinet's encoder combines 1D time-channel separable convolutions following QuartzNet \cite{Kriman2020QuartznetDA} and the SE mechanism \cite{se-DBLP:journals/corr/abs-1709-01507} which has been successfully used in ContextNet \cite{Han2020ContextNetIC} for global context representation. 

Alternatively, ASR systems leveraged on Transformers \cite{transformer-NIPS2017_3f5ee243} and their variants have achieved impressively low word/character error rates (WERs/CERs) in numerous languages during recent years \cite{Miao2020TransformerBasedOC, conformer_gulati20_interspeech, DBLP:journals/corr/abs-2010-11395, Dong2018SpeechTransformerAN,Pham2019VeryDS}. The Transformer architecture includes multi-head self-attention (MHSA) and cross-attention layers that can represent long-distance interactions inside and between sound-text sequences. 

There are works such as Conformer \cite{conformer_gulati20_interspeech} which introduced convolutions into Transformer architecture and achieved impressive WERs/CERs in numerous languages. In Conformer, two macaron-like \cite{macaron-net-lu2019understanding} feed-forward networks (FFN) with half-step residual connections sandwiches the MHSA and convolution modules followed by a post layer normalization \cite{Ba2016LayerN}. 

In this paper, we consider a reverse direction of introducing attention mechanisms into convolution-leading architecture. In particular, we introduce self-attention and cross-attention in a state-of-the-art deep convolutional CTC model, Citrinet, with as minimum modifications as possible. Our work is motivated by the following facts and considerations. First, in order to capture long-range contextual representation, there are 235 convolution layers and 46 linear layers in a typical ``thin and long'' Citrinet model. As will be described in detail in Section \ref{sec:attention-citrinet}, there are 10 convolutional functions stacked in one Jasper block used in Citrinet without internal residual connection. These make the training convergence to be relatively slow and the gradient back-propagation is relatively more difficult without shortcut-style residual connections \cite{resnet-7780459}. Second, kernel sizes from 11 to 39 and further 41 are hard-coded in Citrinet to capture as long range information as possible. This hard setting is less friendly to long and short speech waves: a kernel size of 41 is still not enough for a long wave yet it can be a waste for short waves. Third, current Citrinet is a pure CTC model with a non-autoregressive decoder: one convolutional layer projects from the encoder's output to the vocabulary size. We are interested in boosting Citrinet with cross-attention enhanced decoder architecture. 

In our attention-enhanced Citrinet, we insert a FFN module and a MHSA module in the ``1D conv module'' of one Jasper block and then reduce the repeating time from 5 to 1 to remove 80\% of convolution layers and select 13 from 23 blocks to obtain a fat and short pipeline. We also add a bidirectional Transformer \cite{transformer-NIPS2017_3f5ee243} decoder to ensure Citrinet's training in a non-autoregressive way using a linear combination of CTC loss and attention losses, and inferencing in the same non-autoregressive way. We perform our experiments using the 500-hour Japanese CSJ dataset and the 1600-hour Japanese Magic dataset.



\section{Attention Enhanced Citrinet}\label{sec:attention-citrinet}

\subsection{Citrinet}

\begin{figure}[t]
  \centering
  \includegraphics[width=5.6cm]{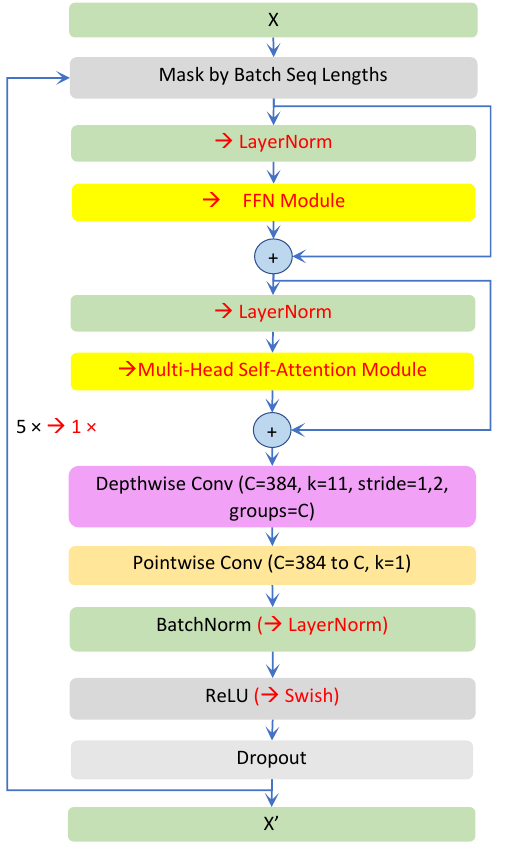}
  \caption{1D time-channel separable convolution module enhanced by FFN and MHSA modules.}
  \label{fig:1d-conv-with-mhs-attention}
\end{figure}

\begin{figure}[t]
  \centering
  \includegraphics[width=6.0cm]{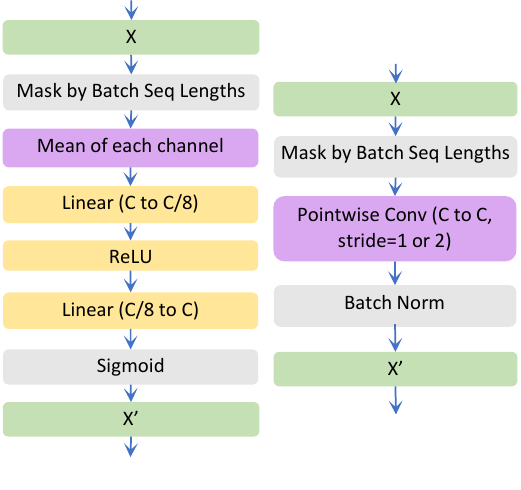}
  \caption{The SE (left) and Res (right) modules used in Citrinet's building block. }
  \label{fig:se_and_res_in_citrinet}
\end{figure}

Figure \ref{fig:jasper_block_with_se_in_citrinet} depicts the ``wave to textual sequence'' pipeline of Citrinet and its major building block. The preprocessing is to extract the 80-dimension FBanks and then perform spectrogram augmentation \cite{spec-aug-2019} on time/frequency dimensions. One (updated) Jasper block \cite{Li2019JasperAE,Kriman2020QuartznetDA}, in the right-hand-side of the figure, contains three modules: (1) one 1D time-channel separable convolution module (Figure \ref{fig:1d-conv-with-mhs-attention}), (2) one SE module \cite{se-DBLP:journals/corr/abs-1709-01507} and in parallel (3) one residual (Res) connection module (Figure \ref{fig:se_and_res_in_citrinet}). This Jasper block can be expressed by the following equations:
\begin{align}
    \tilde{\textbf{X}} & = \text{1D\_Conv}(\textbf{X}) \\ 
    \textbf{X}' & = \text{Res}(\textbf{X}) + \tilde{\textbf{X}} \times \text{SE}(\tilde{\textbf{X}}).
\end{align}

The ``1D conv module'' repeatedly extracts time-channel contextual representations guided by increased kernel sizes. There are two linear layers in the SE module that project from $C$ to $C/8$ and back to $C$ to capture the whole sequence contextual information of each channel. The contextual information is then projected into the $[0, 1]$ range by a sigmoid function to act as \emph{gates} controlling the traffic to the following layers. Thus, the local and global information are extracted by 1D conv and SE modules, respectively. The Res module takes the original tensor $\textbf{X}$ as input and performs a simple $1\times 1$ convolution followed by a batch normalization. The stride for this $1\times 1$ convolution can take values of 1 or 2 to align with the distilling of the three mega blocks to be described hereafter.

The encoder contains 23 Jasper blocks which are assigned to 5 block sets: one prolog block, three mega blocks and finally one epilog block. Prolog block and epilog block both adapt the Jasper block architecture except that the residual module (``Res Module'' in the Figure) is not included. Using the ``1D Conv Module'', prolog block's major task is to transform the input 80-dimension FBanks into $C$ (e.g., 384, 1024) channels and epilog block's major task is to capture global information again with kernel size 41 and then transform the input $C$ channels into the dimension (e.g., 640) similar to the decoder's input. 

There are 6, 7, and 8 Jasper blocks in the three mega blocks, respectively. In each first Jasper block, a stride of 2 is used and consequently the ``time-dimension'' length is halved. Generally, this can be recognized as a \emph{distilling} process in the encoder: output length is $1/2^3$ of input length. In order to capture global context information, the kernel sizes of the mega blocks increase linearly from 11 to 39. In particular, the first mega block's kernel sizes range over 11, 13, ..., to 21; the second mega block takes values from 13, 15, ..., to 25; and the third mega block starts from kernel size of 25 and ends at 39.

The decoder is a one-layer convolutional network that projects from the encoder's output to the vocabulary size. Finally, CTC loss \cite{ctc-10.1145/1143844.1143891} is used during training the architecture. 


\subsection{Enhance 1D Time-channel Convolution by Attention}

Figure \ref{fig:1d-conv-with-mhs-attention} shows the original 1D time-channel separable convolution module and our modifications in red font after arrows. This depth-wise separable network includes (1) a depth-wise CNN with kernel size from 11 to 39 that tries to capture relatively a long range of context and (2) a pointwise CNN that acts like a linear projection with kernel size of 1. Then, batch normalization, ReLU activation and dropout are performed. In the original Jasper block, these repeat 5 times resulting in 10 convolutional functions, 5 batch normalization and 4 ReLU/dropout functions (the final repeat does not include ReLU/dropout). These 10 convolutional layers are stacked together without any residual connections inside them. As shown in Figure \ref{fig:jasper_block_with_se_in_citrinet}, there is only one residual module outside the 1D conv module. These plain nets that simply stack CNN layers (without shortcut-style residual connections in between) exhibit higher training error when the depth increases as illustrated in ResNet \cite{resnet-7780459}.

Motivated by this observation and the successful combination of CNNs with attentions in Conformer \cite{conformer_gulati20_interspeech}, we modify this 1D conv module by (1) setting repeat time from 5 to 1, (2) appending a FFN module and a MHSA module right before the first convolutional layer, (3) changing the batch normalization into layer normalization \cite{Ba2016LayerN} which has been successfully used in Transformer \cite{transformer-NIPS2017_3f5ee243} and (4) replacing ReLU by the Swish \cite{swish-DBLP:journals/corr/abs-1710-05941} activation function which is a self-gated function $f(x) = x \cdot \text{sigmoid}(x)$ and has been successfully used in Conformer \cite{conformer_gulati20_interspeech}.

We reduce 8 convolution functions in one Jasper block. The appending of the FFN and MHSA modules bring 6 linear projection layers, 2 layer normalization and 2 residual connections, making the resulting Citrinet architecture to be fatter and shorter. We count on these FFN+MHSA modules for capturing global interactive information among elements of the input sequence. This works together with the SE module to perform \emph{self-attending} and \emph{self-gating} of global information. 

\subsection{Enhance Decoder by Attention}

We append a 3-block bidirectional Transformer decoder which encodes the textual sequences in left-to-right ($l2r$) and right-to-left ($r2l$) directions and predicts the whole target sequence by one forward step. The input and reference output to the $l2r$ decoder are ``$s, y_1, ..., y_N$'' and ``$y_1, ..., y_N, /s$'' which are taken from a reference label sequence $y_1, ..., y_N$, respectively. Here, $s$ and $/s$ stand for start/end of sequence symbols, respectively. The $r2l$ decoder takes ``$s, y_N, ..., y_1$'' as input and predicts ``$y_1, ..., y_N, /s$''.

The objective is to minimize $\mathcal{L}$, a linear combination of the CTC loss \cite{ctc-10.1145/1143844.1143891} and attention losses (ATT) computed by point-wise KL-divergence \cite{KL-10.1214/aoms/1177729694}:
\begin{equation}
    \mathcal{L} = \lambda_1 * \text{CTC}_{loss} + (1-\lambda_1)*\{\lambda_2 * \text{ATT}_{l2r} + (1-\lambda_2)*\text{ATT}_{r2l}\}
\end{equation}
When we set $\lambda_1=1$, it degenerates to the original Citrinet's loss function. In our attention-enhanced Citrinet variants, we set $\lambda_1=0.3$ and $\lambda_2=0.7$. Label smoothing with $\delta=0.1$ is applied to the attention objective so that the references are discounted by (1-$\delta$).

\section{Experiments}

\subsection{Setup}

Our experiments are performed under an open-source platform, NVIDIA NeMo\footnote{\url{https://github.com/NVIDIA/NeMo}} which is a conversational AI toolkit built for researchers working on ASR, natural language processing, and text-to-speech synthesis with rich pretrained models and complete implementation of Citrinet. We investigate Citrinet and attention-enhanced Citrinet's performances under channel sizes from 256 to 1,048. We release the code\footnote{\url{https://github.com/Xianchao-Wu/nemo_bidecoder}} and pretrained models through NeMo. During training, we select the Novograd optimizer \cite{DBLP:journals/corr/abs-1905-11286-novograd} ($\beta=[0.8, 0.25]$) with a start learning rate of 0.05. The cosine annealing \cite{DBLP:journals/corr/LoshchilovH16a-cosine-annealing} learning rate scheduler with 10K warm-up steps is used. 


We select another open-source ASR platform WeNet\footnote{\url{https://github.com/wenet-e2e/wenet}} \cite{wenet-DBLP:journals/corr/abs-2102-01547} as baseline. WeNet's implementation of the Conformer baseline model consists of a 12-block Conformer encoder ($d_{\text{FFN}}$=2048, $h$=8, $d$=512, CNN$_{\text{kernel}}$=31) and a 3-block bidirectional Transformer decoder ($d_{\text{FFN}}$=2048, $h$=8, $d$=512) which also encodes the textual sequences in $l2r$ and $r2l$ directions.
The models are trained with static batching skills. For decoding, we follow the CTC beam search + Attention-rescoring strategy. All our experiments were performed under NVIDIA DGX-A100 with 8*A100-80GB GPUs.

\subsection{Data}

We evaluate attention-enhanced Citrinet variants on the Japanese ``Corpus of Spontaneous Japanese'' (CSJ) dataset\footnote{\url{https://ccd.ninjal.ac.jp/csj/en/}} and the Japanese Magic dataset\footnote{\url{https://www.Magicdatatech.com/datasets?language=Japanese}}, which respectively consist of 500 and 1,600 hours of labeled waves. The CSJ dataset contains research presentations that are recorded from conference rooms. The Magic dataset is relatively noisier of spoken Japanese with informal Japanese grammar. Both datasets are Tokyo-region pronunciations. We use sentencepiece-bpe \cite{kudo-richardson-2018-sentencepiece} to segment the text sequences and select a vocabulary size of 4,096. Both datasets have 3,000 around single characters and 92\% characters are overlapping. During data preparation, we generate 80-dimension FBank feature vectors with a 25ms window, a 10ms frame stride and dither=1.0. SpecAugment \cite{spec-aug-2019} is adapted with 2 frequency masks ($F$=10) and 2 time masks ($T$=50). Global Cepstral Mean and Variance Normalization \cite{cmvn-10.1016/S0167-6393(98)00033-8} technique is employed to normalize the 80-dimension FBank vectors.

\subsection{Major Results on the CSJ Test Sets}

\begin{table}
  \caption{CERs (\%) of 3 baselines and 5 Citrinet (C) and attention-enhanced Citrinet (Att-C) variants with different channel sizes. All models are trained under CSJ-500h dataset.}
  \label{tab:main_res_compare_csj}
  \centering
  \begin{tabular}{ccccc}
 \toprule
 & 1.73h & 1.82h & 1.23h & Model\\ 
Model & test1 & test2 & test3 & size (M)\\ 
\midrule
Transformer \cite{Li2019ImprovingTS} & 7.6 & 6.1 & 6.3 & 36\\ 
Espnet+LM & 6.5 & 4.6 & 5.1 & NA\\ 
Conformer \cite{conformer_gulati20_interspeech}  & 6.97 & 4.65 & 5.29 & 135.1\\
\midrule
C-256 & 8.14  & 5.79  & 6.54  & 12.2\\ 
C-384 & \textbf{7.28}  & \textbf{4.81}  & \textbf{5.44} & 22.7\\ 
C-512 & 7.54  & 5.45  & 6.01  & 38.0\\ 
C-768 & 7.37  & 5.33  & 5.71  & 85.0\\ 
C-1024 & 7.58  & 5.76  & 5.87  & 141.6\\ 

\midrule
Att-C-256 & 7.10  & 4.75  & 5.79  & 19.9 \\ 
Att-C-384 & \textbf{6.27}  & \textbf{4.16}  & \textbf{5.01}  & 40.3 \\ 
Att-C-512 & 6.99 &	4.94 &	5.46   & 68.6 \\ 
Att-C-768 & 7.23 &	4.87 &	5.18   & 123.5 \\ 
Att-C-1024 & 7.23 &	5.49 &	5.55  & 201.4 \\ 
\bottomrule
  \end{tabular}
\end{table}

Table \ref{tab:main_res_compare_csj} lists the CERs (\%) of 5 Citrinet and attention-enhanced Citrinet variants and 3 baselines. The first baseline is a Transformer with enhanced information from speakers and speech \cite{Li2019ImprovingTS}. Parameters were shared among layers of encoder and decoder blocks in Transformer. Speaker information (gender, age, education-level, speaker ID) and speech utterance properties (duration of short and long, topic of the lecture) were used to augment the training data. The second baseline is Espnet's implementation of deep VGG-BLSTM \cite{inproceedings-vgglstm} with CTC joint decoding and LM rescoring\footnote{\url{https://github.com/espnet/espnet/blob/master/egs/csj/asr1/RESULTS.md}}. Note that only this baseline uses an external language model among all the 13 models. The third baseline is Conformer-Large \cite{conformer_gulati20_interspeech} with 135.1M parameters. 

Since convolutions play a leading position in Citrinet, we take channel sizes of 256, 384, 512, 768, and 1024 to investigate if enlarging the model size yields lower CERs. From Table \ref{tab:main_res_compare_csj}, we observe that C-384 performs the best and larger channel sizes hardly bring significant reduction of CERs. Similar tendency happens when we attach an attention mechanism to Citrinet. This does not strictly align with Citrinet's performance on the famous LibriSpeech dataset. We make a detailed analysis of the length of the waves and find that 76.6\% of LibriSpeech's waves are longer than 10 seconds, yet 97.8\% of CSJ's waves are less than 5 seconds. This reflects that large channel sizes in CNN layers are less helpful to ASR corpus with short waves.

In addition, we have the following observations. First, Conformer performs well among the baselines, even with a relatively large model size. Conformer's performance is comparable with Espnet where an external LM is employed for rescoring.

Second, under all the 5 channel sizes, attention-enhanced Citrinet (Att-C) performs better than their Citrinet counterparts, with averagely 0.94\%, 0.70\%, 0.54\%, 0.38\%, and 0.31\% CER reducing, respectively. These reflect that attentions do help Citrinet capture global contextual representations. The model sizes are nearly doubled by adding FFN+MHSA modules and $l2r, r2l$ decoders. We set $h$=8, $d$=256 for Att-C-256, $d$=384 for Att-C-384, and $d$=512 for the other three variants, and $d_{FFN}$=4$d$. With these modifications, Att-C converges faster than Citrinet and only requires 80\% of the total training time.

Third, the best Att-C-384 also outperformed Conformer by reducing 0.49\% (relatively 8.7\%) CER scores on average with 29.8\% model size (40.3M vs. 135.1M). Conformer and Att-C-384 are set the same number of training epochs. Training one epoch for Conformer costs 10min yet Att-C-384/C-384's are respectively 4min/5min. We thus only require 40\% of Conformer's training time in total with the same epoch number.


\subsection{Major Results on the CSJ+Magic Datasets}

\begin{table}
  \caption{CERs (\%) of 3 baselines and 3 Citrinet (C) and attention-enhanced Citrinet (Att-C) variants with different channel sizes. All models are trained under the combined CSJ-500h + Magic-1600h dataset.}
  \label{tab:main_res_compare_csj_Magic}
  \centering
  \begin{tabular}{ccccc}
 \toprule
 & 1.73h & 1.82h & 1.23h & 2.01h\\ 
Model & test1 & test2 & test3 & Magic-test\\ 
\midrule
Conformer \cite{conformer_gulati20_interspeech} & 11.22  & 6.35  & 8.78  & 11.73 \\ 
\midrule
C-256 & 13.68  & 7.78  & 7.29  & 12.65 \\ 
C-384 & 8.51  & 5.80  & 8.36  & 10.56 \\ 
C-512 & 12.39  & 7.25  & 7.72  & 13.98 \\ 
\hline
Att-C-256 & 12.02  & 7.22  & 7.54  & 10.80 \\ 
Att-C-384 & \textbf{7.94}  & \textbf{5.62}  & \textbf{6.92}  & \textbf{9.13} \\ 
Att-C-512 & 11.53  & 7.21  & 7.60  & 11.98 \\ 
\bottomrule
  \end{tabular}
\end{table}

Considering the fact that the Japanese CSJ dataset only contains 500 hours of training data and most of its waves are less than 5 seconds, we append a 1600-hour Japanese Magic dataset of spoken Japanese contents which has 67.64\% wave files that exceed 10 seconds. The CSJ dataset contains presentations of research conferences. This brings a challenge of domain adaptation. We are interested in disclosing if the Conformer baseline and Citrinet variants can keep their performance. From the 1600-hour Magic dataset, we randomly pick out 20 speaker ids and take half as the development set and the remaining half as the test set. Both sets contain 2 hours and there is no overlapping of the speaker ids in the training data anymore. We combine the two training datasets and retrain the sentencepiece-bpe tokenizer with a same vocabulary size of 4,096. 

Table \ref{tab:main_res_compare_csj_Magic} lists the CERs of three Citrinet and attention-enhanced Citrinet variants and the Conformer baseline with the same configuration as Table \ref{tab:main_res_compare_csj}. We choose channel sizes 256, 384, and 512. From Table \ref{tab:main_res_compare_csj_Magic}, we observe that C-384 again performs the best and larger channel sizes do not bring reduction of CERs: C-256 and C-512's performances are comparable. Similar tendency happens in attention-enhanced Citrinet. 

We have the following observations. First, CSJ's three test sets' CERs become worse when different domain dataset is merged into the relatively clean CSJ dataset. Second, the original Citrinet performs better than the Conformer, with 1.21\% (relatively 12.7\%) CER decreasing on average of the four test sets. This reflects that a deeper convolutional model trained using large-scale datasets is still robust. Finally, attention-enhanced Citrinet outperforms Citrinet of all the four test sets with a 0.91\% (relatively 10.9\%) CER reduction on average. 

\subsection{Impact of Individual Modules}

\begin{table}
  \caption{CERs (\%) of variants when deleting individual modules on the CSJ test sets.}
  \label{tab:main_res_ab}
  \centering
  \begin{tabular}{lcccc}
 \toprule
Model & test1 & test2 & test3\\ 
\midrule
C-384 & 7.28  & 4.81  & 5.44 \\ 
\hline
Att-C-384 & \textbf{6.27}  & \textbf{4.16}  & \textbf{5.01} \\
\ \ without bi-decoder &  7.08  & 4.51  & 5.24 \\
\ \ without FFN &  6.95  & 4.65  & 5.25 \\
\ \ layer-norm $\rightarrow$ batch-norm &  6.58  & 4.46  & 5.21 \\ 
\ \ Swish $\rightarrow$ ReLU &  6.35  & 4.29  & 5.23 \\ 

\bottomrule
  \end{tabular}
\end{table}

We perform an ablation study of the individual ideas in the attention-enhanced Citrinet. Table \ref{tab:main_res_ab} shows the CER results when independently removing bi-decoder, dropping FFN modules, replacing layer normalization with batch normalization and changing Swish activation function back to ReLU. Bi-decoder and FFN are relatively important among the four ideas.

\subsection{Ablation Study of Jasper Block Numbers}

We finally investigate the number of total Jasper blocks with a value set of \{8, 9, 10, 11, 12, 13, 15, 17, 20\}. We averagely prune ending Jasper blocks in the 3 mega block set (Figure \ref{fig:1d-conv-with-mhs-attention}). The variants achieve stable comparable CERs with a standard deviation of 0.3\% except 9 and 12 with +2\% CERs. We thus select the first stable block number of 13.

\section{Conclusions}



In this paper, we introduce attention mechanisms into a convolution-leading architecture, Citrinet, for end-to-end non-autoregressive Japanese ASR. We keep Citrinet's architecture of using 1D time-channel separable convolutions, SE and residual modules and append FFN and MHSA before each 1D time-channel convolution. We reduce 8 convolution layers in each attention-enhanced Citrinet block and only utilize 13 such blocks. Experiments on the Japanese CSJ-500h and Magic-1600h datasets show that the attention-Citrinet requires less layers of blocks and converges faster with lower character error rates than Citrinet with 80\% training time and Conformer \cite{conformer_gulati20_interspeech} with 40\% training time and 29.8\% model size. Future work includes using linear attentions \cite{9423033-linear-attention, Li2021EfficientCS_linear_conformer} and adaptive span attentions \cite{Chang2020EndtoEndAW} for building more efficient Citrinet. 



\bibliographystyle{IEEEtran}

\bibliography{mybib}


\end{document}